\documentclass[conference]{IEEEtran}
\IEEEoverridecommandlockouts

\usepackage{cite}
\usepackage{amsmath,amssymb,amsfonts}

\usepackage{graphicx,subfigure}

\usepackage{lipsum, hyperref}

\begin{document}
\title{Model Exploration with Cost-Aware Learning}

\author{N. R. Stillman, I. Balazs, S. Hauert}

\maketitle

\begin{abstract}
We present an extension to active learning routines in which non-constant costs are explicitly considered. This work considers both known and unknown costs and introduces the term $\epsilon$-frugal for learners that do not only consider minimizing total costs but are also able to explore high cost regions of the sample space. We demonstrate our extension on a well-known machine learning dataset and find that out $\epsilon$-frugal learners outperform both learners with known costs and random sampling. 

\end{abstract}

\maketitle

\section{Introduction}

Recent advances in machine learning have translated into a wide range of options for the exploration and analysis of data. Furthermore, increased computing power along with advances in experiment design (such as micro- and bio-fabrication procedures) have meant that data generation is both cheaper and faster to generate. This allows for huge data to be generated from both simulations and experiment, providing  opportunities for new understanding of complex models. However, the efficient exploration of these models can be made especially difficult exactly due to the massive volume of data. 

Instead of random or exhaustive searching of the parameter space, autonomous methods for model exploration reduce the number of costly experiments by focusing on areas of analysis in which there is highest uncertainty in the results of the model \cite{Settles2012, Settles2010}. Such methods, known collectively as active learning, extend the power of current machine learning tools to experiment design and have been well applied in fields such as drug discovery, natural language processing and the experiment design \cite{Murphy2011, Fang2017, King2004}. However, the majority of these methods assume that costs are equal for all experiment or simulation results and hence, try to minimize the total number of queries \cite{Settles2008}. Yet, for many experiments and simulations, queries largely vary in costs in ways that are not straightforward. 

In this paper, we present a method for model exploration with cost-aware learning. We build on the popular and well-used machine learning toolbox of scikit learn \cite{Pedregosa2011}. However, the learning algorithms can be interchanged with other custom-build routines. The learning occurs both in training the meta-model of the simulation as well as in learning the costs associated with running a simulation where that data is unknown. 

This paper is set out as follows. In Section II, we outline our active learning routine and how it accounts for costs. We demonstrate our model exploration tool on the well-known breast cancer Wisconsin (diagnostic) dataset  from the UCI Machine Learning repository \cite{Dua2019 }. One of the key benefits of using our method is the accounting for costs such as simulation time that are not necessarily uniform across all parameter samples. Considerations of costs are controlled by a $\epsilon$-frugal which we discuss in more detail in Section III. Finally, in Section IV, we describe future extensions to our tool, before making out concluding comments.

\section{Overview of Active Learning}

Active learning is a form of semi-supervised learning, in that it requires labeled instances in order to train a model. However, it differs from many conventional forms of supervised learning in that it seeks to minimize the number of labeled instances used, either because availability of data is limited or because data is costly to obtain. Here, we assume that all data are available in a large unlabeled pool and that labeling of data occurs by querying an `Oracle’. We refer the reader to \cite{Settles2011} for further information on active learning implementations. In practice, this oracle may represent either a simulation or an experiment run and, hence, will have costs associated with it. In this work, we include considerations of these costs explicitly. 

We assume that the active learner will sample from a pool of size $N$ of unlabeled data, $\{x_i\} \in \mathcal{U}^N$ for $i=1\ldots N$. In this work, we demonstrate our active learning framework on a binary classification problem, such that the trained model is a binary-valued function $f : \mathcal{U}^N \rightarrow \{0,1\}$ with some probability of success $P_f(x_i)$. The true label of data is obtained by querying the Oracle, which maps unlabeled data, $x_i$, to a labeled data, $y_i$, i.e. $O(x_i)$ is equal to $f(x_i)$ but where $P_f(x_i) = 1$ for all $x_i$. 

Prior to initializing the active learner, a subset of $t_0$ unlabeled data are labeled by the Oracle, $\{y_{t_0}\} \in \mathcal{L}^{t_0}$. At each subsequent iteration, $t_n$ queries are then taken from $\mathcal{U}$ to be queried. This results in a set of labeled data, $\{y_{t_0}\} \in \mathcal{L}^M$ for $j=1\ldots M$ and where $M = \sum_{n=1}^Tt_n$. Here, $T$ represents the total number of iterations and $t_n$ represents the number of samples queried for each iteration. The initial sample $t_0$ is generally chosen to be larger than $t_n$ to ensure sufficient initial model accuracy and $t_n$ is constant across iterations. At each iteration, the queries are added to the labeled set and removed from the unlabeled set such that there is a constant increase in the training data available, i.e. $\mathcal{L} = \mathcal{L}^{t_0} \subset \mathcal{L}^{t_1} \ldots \subset \mathcal{L}^{t_T}$. 

As discussed, the aim of active learning is to train a model using the smallest number of samples but with high model accuracy. That is we want to minimize $m$ whilst maximizing $P_f(x_i)$. In order to maximize $P_f(x_i)$, we require a sufficient metric to rank available unlabeled samples. Given this metric, we then choose those that have the lowest probability of success, i.e. those that the model is least certain on. This metric underpins the method for selecting queries to pass to the Oracle and are known in the literature as `query selection' measures \cite{Huang2014}. 

For active learning, there have been many such measures proposed, such as the Kullback-Leibler (KL) divergence, the Fisher information ratio, and the overall information density \cite{Kullback1951, Zhang2000, Settles2008}. For more information on query selection, see \cite{Settles2010}. In this work, we train a single model and focus on the well-known uncertainty sampling methods of least-certain sampling, margin-based sampling and entropy-based sampling. Future work will introduce ensemble or committee query selection, which is relevant for when multiple models are trained simultaneously.

One of the most simple methods for query selection is to select each sample that has the lowest certainty (i.e. the lowest probability of correct labeling) \cite{Culotta2005}. We refer to this as least-certain sampling (LC),  
\begin{equation}
\phi^{LC}(x_i) = 1 - P_f(y^*| x_i):
\end{equation}
where $y^*$ is the most likely label for unlabeled data $x_i \in \mathcal{U}$.
One of the downsides of using least-certain is that it uses only one of probabilities of prediction. Traditional margin-based sampling increases the number of datapoints by one by calculating the difference between the two most likely labels \cite{Scheffer2001}, 
\begin{equation}
\phi^{MB}(x_i) = -(P_f(y_1^*| x_i) - P_f(y_2^*| x_i)).
\end{equation}
However, when implemented, we find that the distance between both labels to give higher accuracy in general. Hence, we instead implement margin-based sampling as
\begin{equation}
\phi^{MB*}(x_i) = |P_f(y_1^*| x_i) - P_f(y_2^*| x_i)|,
\end{equation}
where $||$ refers to the Euclidean distance between labels.
Finally, entropy-based sampling uses all data to rank the data for selection. For a random variable, $X$, entropy describes the amount of information required to describe the distribution of outcomes of $X$ \cite{Shannon1948}. This is calculated as, 
\begin{equation}
\phi^{EB}(x_i) = -\sum_{l}^{\mathrm{labels}} P_f(y_l| x_i)\log\left(P_f(y_l| x_i)\right),
\end{equation}
where labels refers to the possible labels for unlabeled data, which is two for our binary classification problem. 

Finally, the training of a model involves the trade-off between model exploration and exploitation \cite{sutton2018}. Exploitation refers to a model using it's knowledge (predictive power) to inform it's decision. In the active learning framework, this means that a learner will exploit their knowledge when unlabeled data that is selected to be queried is that which most minimizes model uncertainty. Alternatively, exploration is when a model considers other options which appear sub-optimal. Within the active learning framework, these are those which the model has a high probability of predicting/classifying correctly. An active learner is said to be \textit{greedy} when it is entirely exploitative and $\epsilon$-greedy when it chooses to explore (choose samples that do not maximize accuracy) with probability $\epsilon_g$. 

This summarizes the basic framework for an active learner. Multiple extensions have been investigated, including ensembles or committees of models, alternative query selection methods, and the influence of $\epsilon$-greedy approaches \cite{Melville2004, burbidge2007active, Settles2008, gervasio2005active, Borkowski2020}. However, few active learning frameworks also explicitly consider the influence of costs on learning. We introduce cost-awareness in the next section.  

\begin{figure}[!b]
	\centering
	\includegraphics[width=\linewidth]{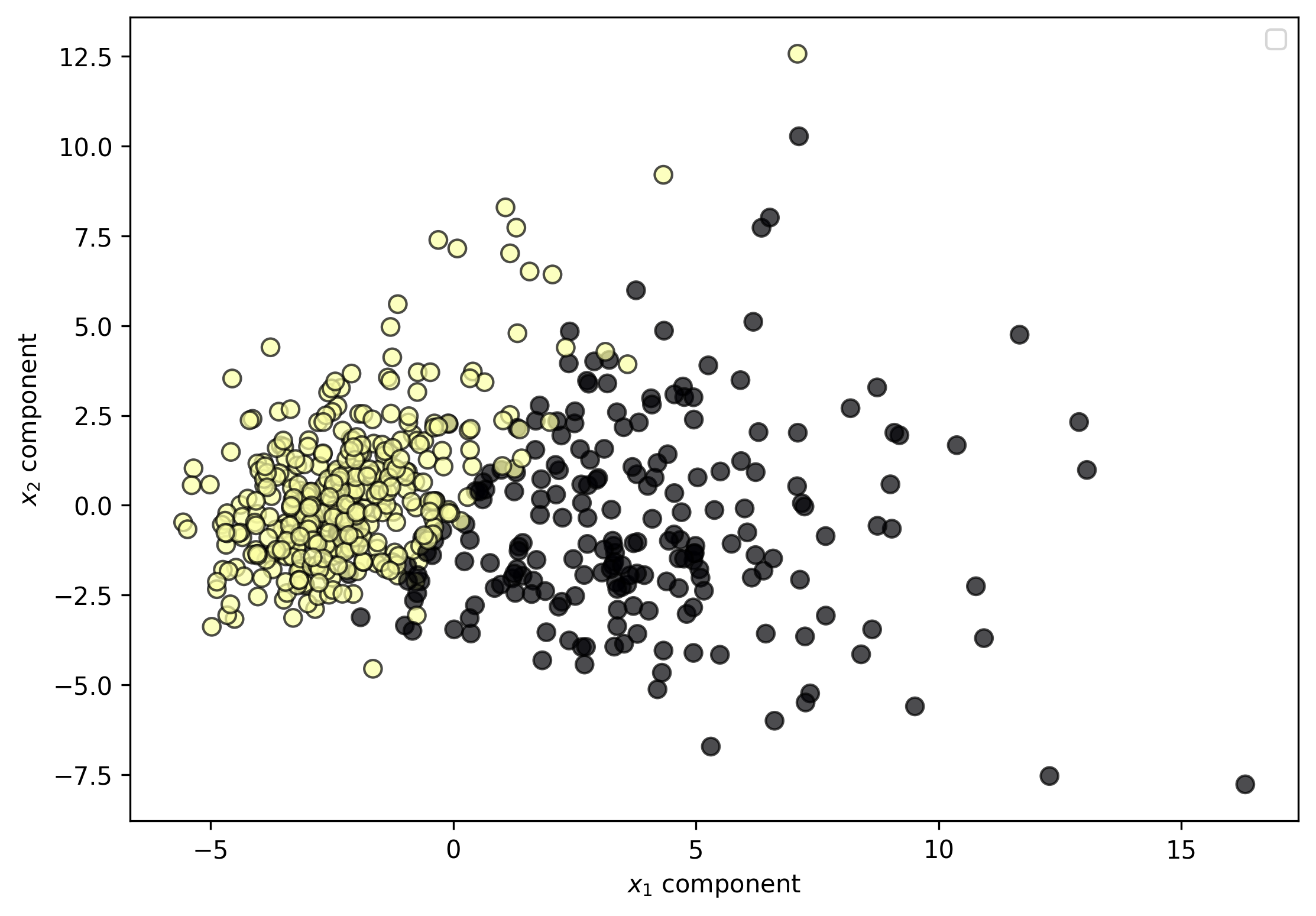}
	\caption{First two components of PCA decomposition of the breast cancer Wisconsin (diagnostic) dataset, showing clustering around the two binary classifications.}
	\label{fig:data}
\end{figure}

\section{Introducing Cost-Awareness into Learning}

The majority of active learning routines assume that the costs to query a sample are equal across data, such that the active learning seeks to minimize the total number of samples queried. In practice, however, costs vary across data or may not be known. This can be due to the expense of lab-time or equipment required to run experiments or the computing power required to run complex simulations. Given this, the cost of labeling queries brings an additional consideration to minimizing sample numbers where both the increase to model accuracy as well as the reduction of costs are combined in the selection process.

To balance this, we introduce cost-awareness to the standard active learning framework. This involves three additions to the overview above. First, if costs are known when data is ranked for querying, we adapt the ranking stage to account for costings. For both least-confidence and entropy-based querying, we follow the standard method of dividing all queries by the costs. 

Secondly, as with considerations of the trade-off between model exploitation and exploration, sub-optimum accuracy can be achieved by focusing exclusively on minimizing costs. We refer to these active learners as {\it frugal}. To address the drop in accuracy, we introduce the concept of $\epsilon$-frugality to our active learning framework; an active learner chooses a costly sample with probability $\epsilon_f$. 

Finally, query costs can be unknown at run-time (for example, where humans are included in the labeling process). To account for this, our active learner can regress on costs within the labeled set of data, $\mathcal{L}$ and use these to predict future costs. As we describe below, inaccuracies within the regression process reduces the probability of choosing the sample that minimizes costs and can lead to a similar effect as with $\epsilon$-frugal learners and can result in better accuracy models under certain situations.

\begin{figure}[!t]
	\centering    
	\subfigure[Initial certainty]{\label{fig:initalcertainty}\includegraphics[width=\linewidth]{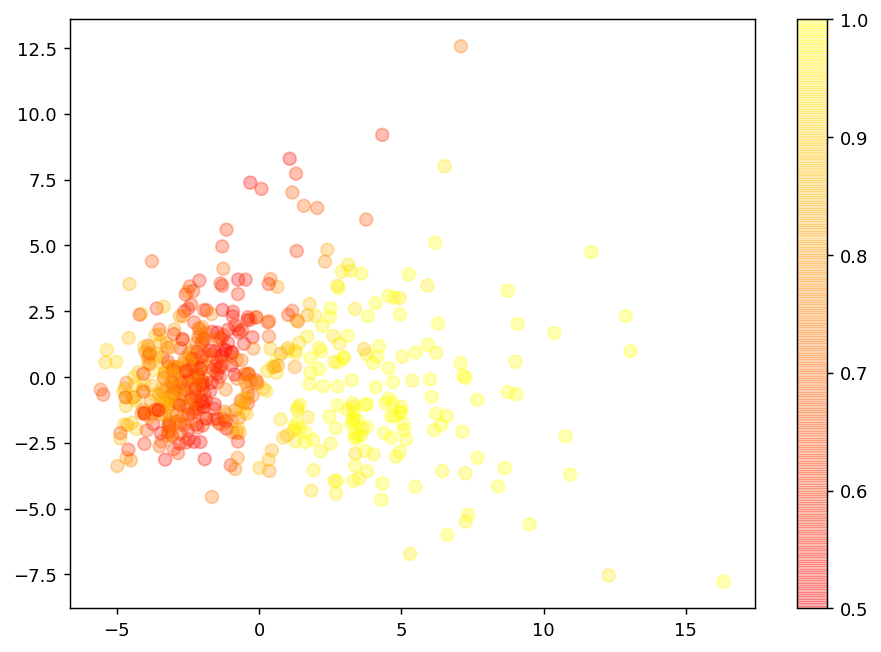}}
	\subfigure[Certainty after Active Learning]{\label{fig:activecertainty}\includegraphics[width=\linewidth]{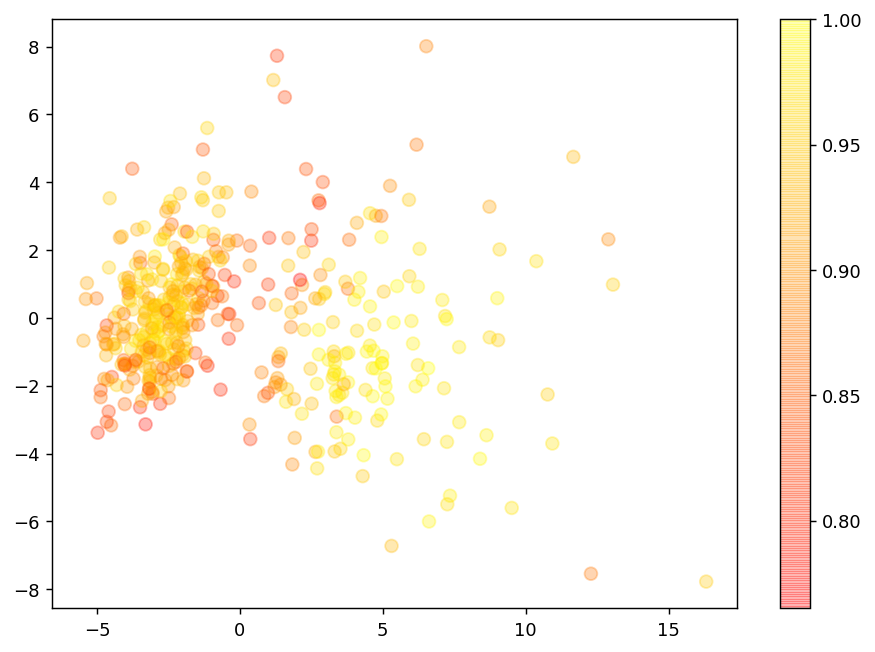}}
	\caption{Demonstration of how points selected by the Active Learner are those that have highest uncertainty, where (a) shows the initial certainty of a Random Forest classifier trained on 5\% of the data (initial accuracy = 0.81) while (b) shows the subsequent certainty of a Random Forest classifier after 50 iterations of the Active Learner, and where 2 datapoints were selected to minimize uncertainty for each iteration. Final accuracy for the active learner is 0.99. Note, the colorbar-scale in (a) and (b) differ.}
	\label{fig:certainty}
\end{figure}

\section{Example of Cost-Aware Learning}

In order to demonstrate our method for cost-aware learning, we use the well-known and widely analyzed breast cancer Wisconsin (diagnostic) dataset. This multivariate dataset consists of 569 individual elements, each with 10 real-valued features and where each datapoint is classified as either malignant or benign. In  \autoref{fig:data}, we show the first two eigenvectors of the data set following a principal component analysis (PCA), where we label the data based on their classification. This shows the data as being clearly segmented into two clusters and provides a useful tool for demonstrating our cost-aware learning method.

\begin{figure}
	\centering
	\includegraphics[width=\linewidth]{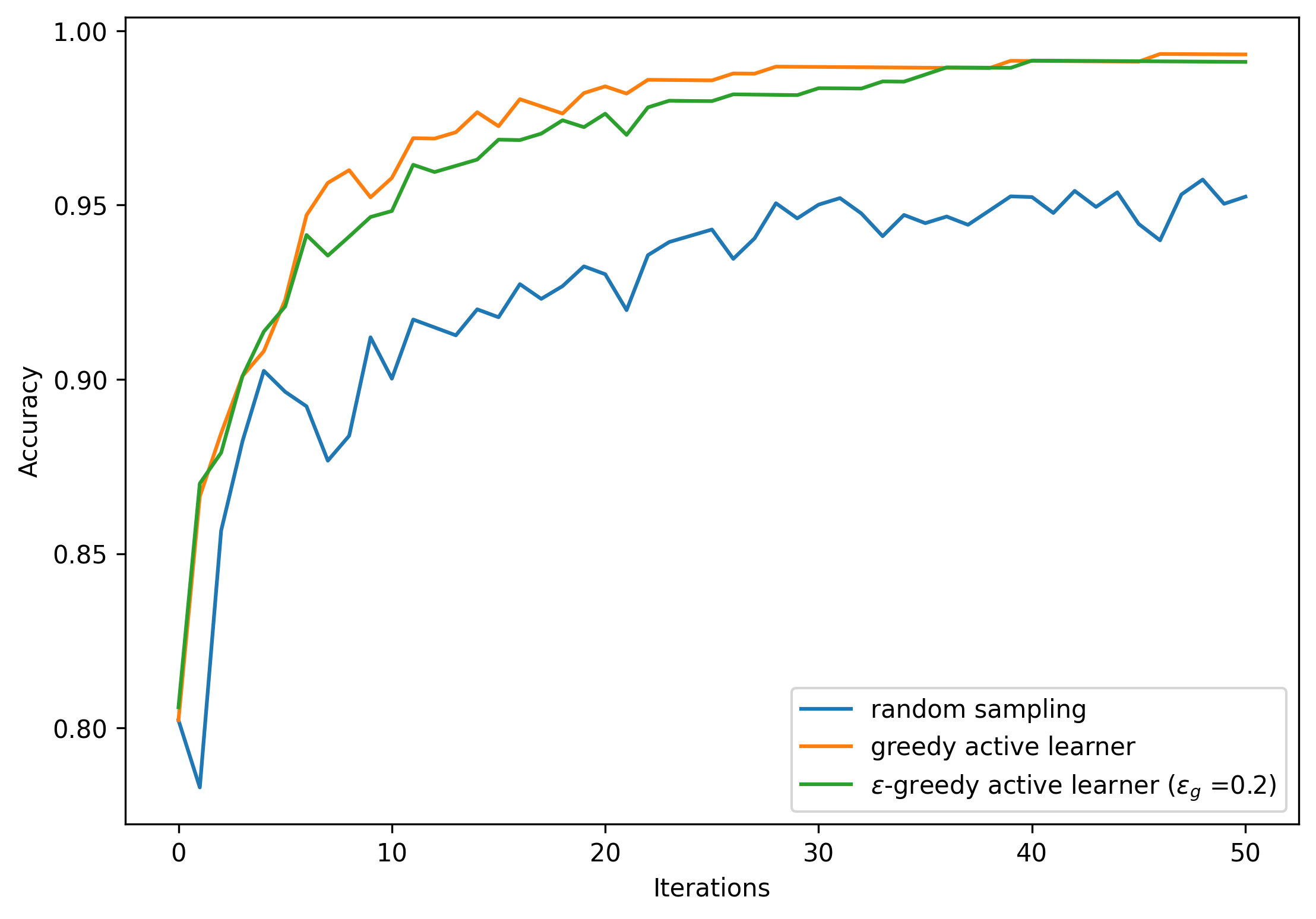}
	\caption{Comparison of the accuracy of training the model without using an active learner (random sampling), using a pure greedy active learner and an $\epsilon$-greedy active learner (for $\epsilon$ = 0.2). Query selection is entropy-based for all active learners.}
	\label{fig:greedyeps}
\end{figure}

As this is a classification problem, we choose a random forest classifier as our model. We use the popular scikit-learn python package to train the model, with 200 estimators and 5 maximum features. Using only 5\% of the data, the classifier is able to correctly classify 81\% of the data. In \autoref{fig:certainty}, we show (a) the initial certainty of the random forest classifier and (b) the certainty after using our active learner. Here, we initialize our active learner with the same 5\% of the data used to initially train the model, run the active learner for 50 iterations, and choose 2 data for querying during each iteration. We use an entropy-based query selection method throughout. The active learner used in \autoref{fig:certainty} is greedy, in that the learner samples exclusively based on whether a sample is thought to increase model accuracy. \autoref{fig:certainty}(b) shows the result of this active learner, where all points with certainty less than 0.75 are removed after 50 iterations. This results in a model with 99.1\% accuracy. 

In \autoref{fig:greedyeps}, we show the learning curves for the greedy active learner, an $\epsilon$-greedy learner (where $\epsilon_g$=0.2), and the equivalent random sampling of the data. We see that all three sampling methods cause an increase in accuracy. However, both active learners quickly rise above 95\% accuracy (after approximately 10 iterations, or 20 samples) and converge to 99\% accuracy by the end of the iterations. The random sampling fails to rise above 95\% accuracy even after the 50 iterations have passed. 

\begin{figure}[!t]
	\centering
	\includegraphics[width=\linewidth]{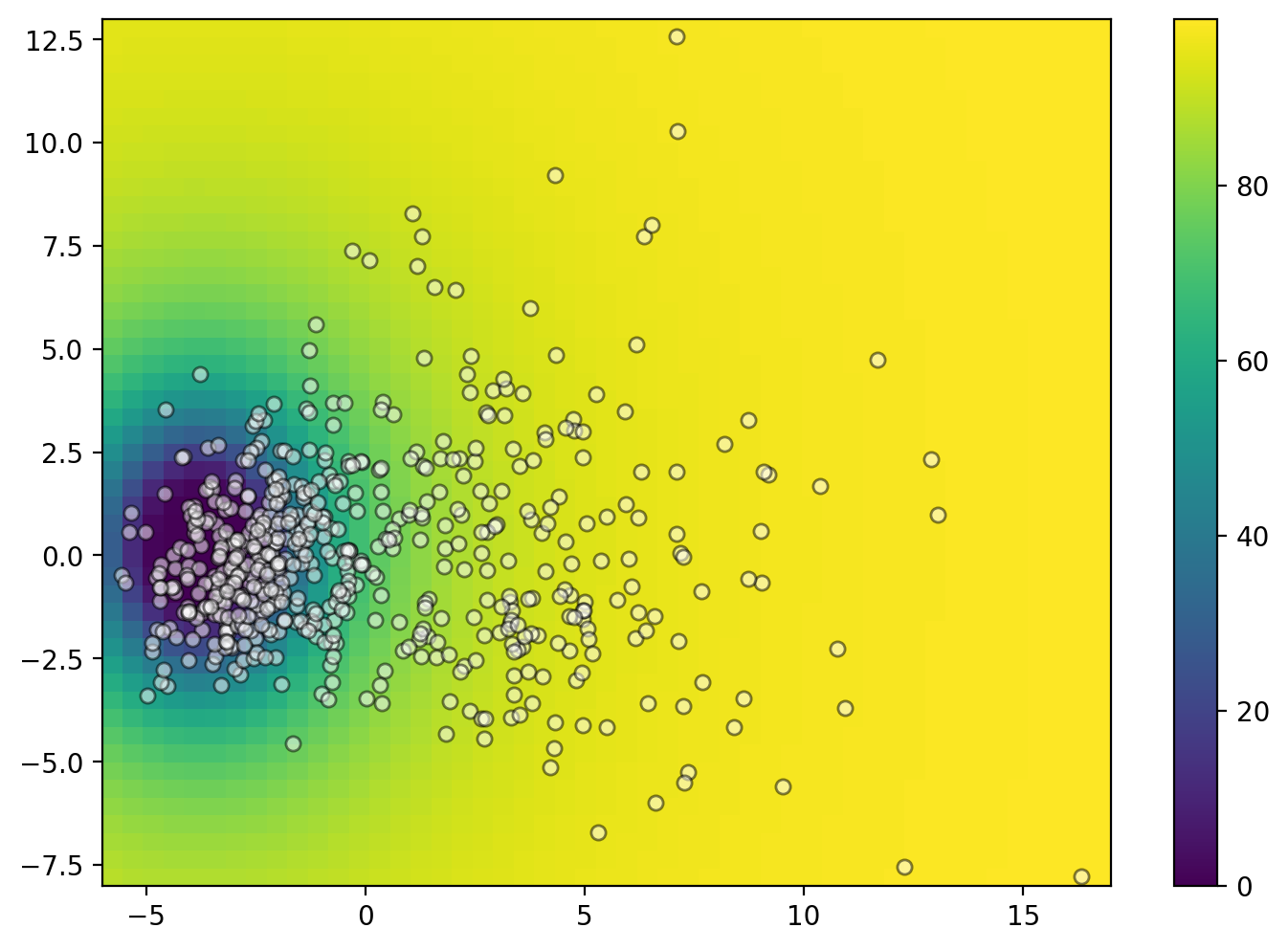}
	\caption{Additional costs imposed on the dataset, where dark (blue) represents low costs and bright (yellow) represents high costs.}
	\label{fig:costs}
\end{figure}

\begin{figure}[t]
	\centering
	\includegraphics[width=\linewidth]{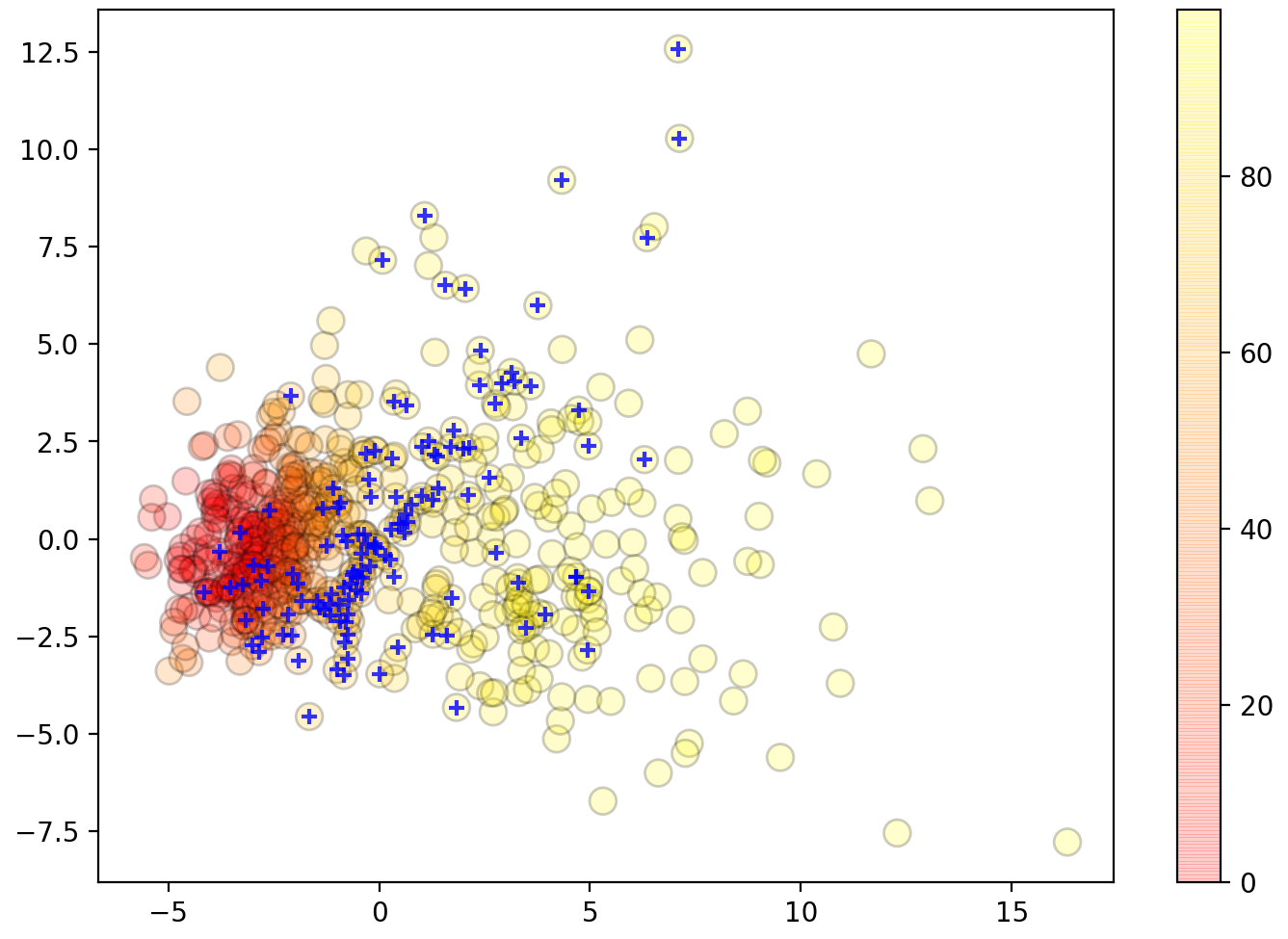}
	\caption{Final sampling after 50 iterations of the Active Learner that does not account for costs. The majority of points selected are at the classification boundary where results were least confident but which have high costs. Query selection is entropy-based.}
	\label{fig:costlysampling}
\end{figure}

So far, all query selections have been made independent of costs. In order to highlight how costs alter the active learning process, we introduce synthetic costs to the data, shown in \autoref{fig:costs}. This is an arbitrary costing that gives low costs that are slightly off-centered to one of the classification clusters and high costs to all other. Importantly, the boundary between both clusters (where uncertainty is high), has a higher cost than those near the center of one of the clusters. Shown in \autoref{fig:costlysampling}, where colour now indicates costs (rather than uncertainty), we see that the majority of points selected are those that are costly to query. We next introduce costs to the learning process in order to address this. 

\section{$\epsilon$-frugal}

As with greedy learning, the exploitation of costs can cause sub-optimal accuracy. This is especially true for learning in a costly environment when regions of highest benefit to model accuracy sit within regions of high costs. By focusing only on low-cost queries, long waits will be observed as the learner seeks to exhaust all low cost queries before moving on to those that are more costly but more beneficial to the learner. However, the cumulative cost of exhausting all low-costs queries can result in larger overall costs than if the learner were to sample some high-cost samples with low probability, causing it to explore costly regions. This is the motivation behind introducing $\epsilon$-frugal learning.

Interestingly, not all forms of costly learning will result in the same cost-rank relationship. This is important as it gives an additional choice to the design of the active learner based on the problem-specific structure of the cost-certainty space. For example, one can imagine situations where costs are normally distributed throughout the sample space such that high costs should be sampled relatively infrequently as other samples are available nearby with lower costs. Alternatively, as with \autoref{fig:costs}, costs may be densely centered around certain regions. In this case, especially where costs overlap with classification boundaries for example, it is important to sample costly samples with relatively high frequency. While our notion of $\epsilon$-frugal learning addresses this by allowing the exploration of costly regions to be adjusted, we observe that the query selection measure can also impact the structure of the sample cost-rank relationship.

\begin{figure}[!t]
	\centering
	\subfigure[Entropy-based query selection  
	]{\label{fig:entropycosts}\includegraphics[width=0.75\linewidth]{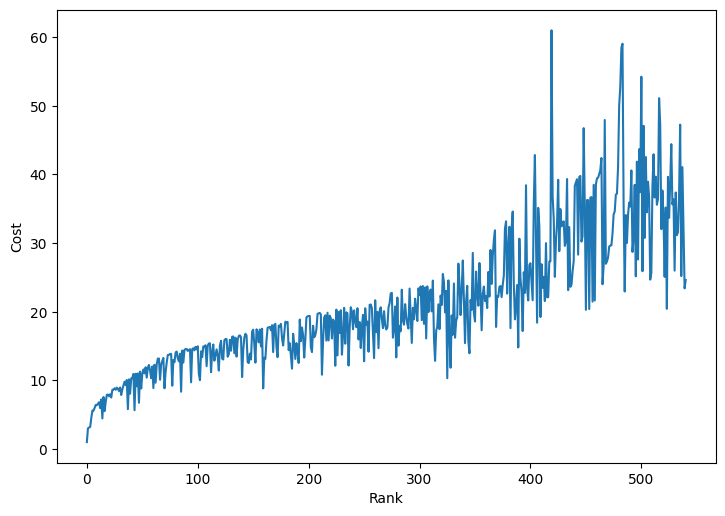}}
	\subfigure[Margin-based query selection  
	]{\label{fig:margincosts}\includegraphics[width=0.75\linewidth]{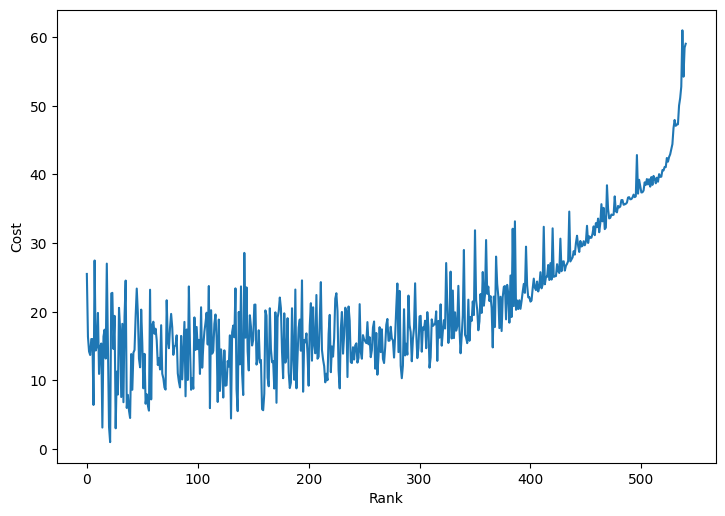}}
	\caption{Comparison of two query selection certainty measure, (a) entropy-based and (b) margin-based. Entropy-based ranks fewer datapoints as low costs than margin-based (where low rank indicates higher probability of selection).}
	\label{fig:measures}
\end{figure}

In \autoref{fig:measures}, we compare how entropy-based certainty measures and margin-based certainty measures relate the rank of data to costs. In entropy-based certainty measures, there are fewer datapoints centered around low costs and low rank (where low rank indicates higher probability of selection). This means that data with higher costs are selected more frequently, when $\epsilon_f$ is non-zero or where the number of iterations of learning is high enough to sample more of the data.

 On the other hand, margin-based certainty measures have many low cost data with low rank. Overall, this means that there is a higher cost for sampling data (where noise around the lower ranked data is caused by the fluctuations in the query selection measure) but data with highest costs are much less likely to be sampled. Hence, margin-based query selection measures may be more appropriate for scenarios where few very high costs are present while entropy-based query selection measures may be more appropriate for where there are many medium to high costs and where they are centered around potential regions of uncertainty. This is the situation we face in cost structure shown in\autoref{fig:costs}, hence we focus on entropy-based selections.

\begin{figure}[t]
	\centering
	\includegraphics[width=\linewidth]{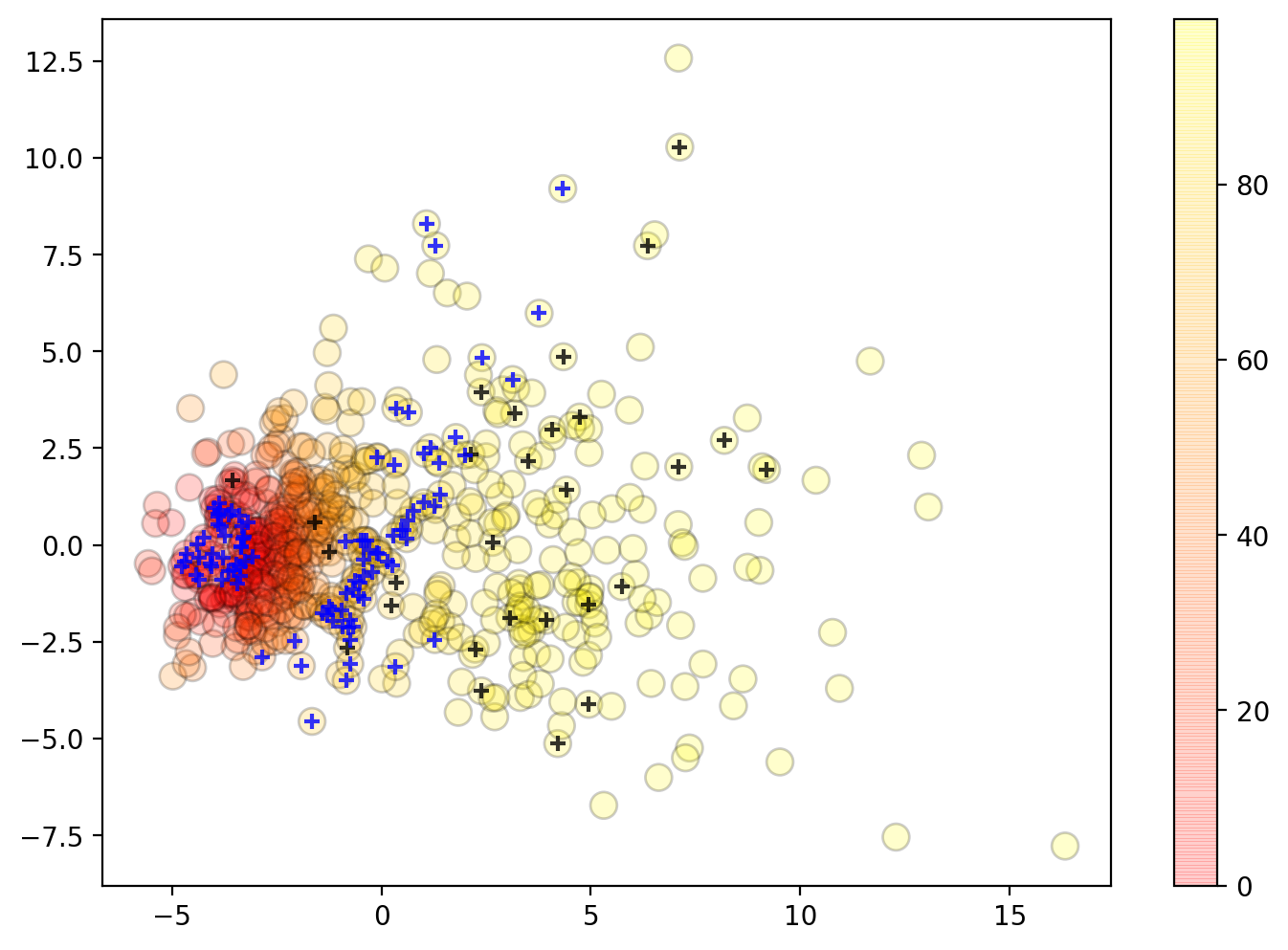}
	\caption{Final sampling after 50 iterations of $\epsilon$-frugal active learner ($\epsilon_f$=0.2). End accuracy is 0.993 but with reduced overall costs (see Figure 8). Query selection is entropy-based.}
	\label{fig:unknowncostlysampling}
\end{figure}

In \autoref{fig:unknowncostlysampling}, we demonstrate the final selection of sampled points from an $\epsilon$-frugal active learner ($\epsilon_f = 0.2$). We find that, while there are significant numbers of the same points sampled between classification clusters, as with \autoref{fig:costlysampling}, there are also many points sampled around regions of lowest cost. The overall accuracy of the model after using the $\epsilon$-frugal learner is 99.3\%. 

Finally, to highlight situations where costs are unknown at query stage, we also apply an active learner that also predicts sample costs to this data. The prediction of the costs, given queried data, is a regression problem and we use Gaussian processes (GP) to solve this. We use the standard implementation of Gaussian processes provided by the scikit-learn package, with a constant kernel with constant value of 1 and with upper and lower bounds for this value between 0.001 and 1,000. For the GP regression parameters, we select $\alpha =  0.1$ with 10 restarts for the optimizer and where target values have been normalized. 

\begin{figure}
	\centering
	\includegraphics[width=\linewidth]{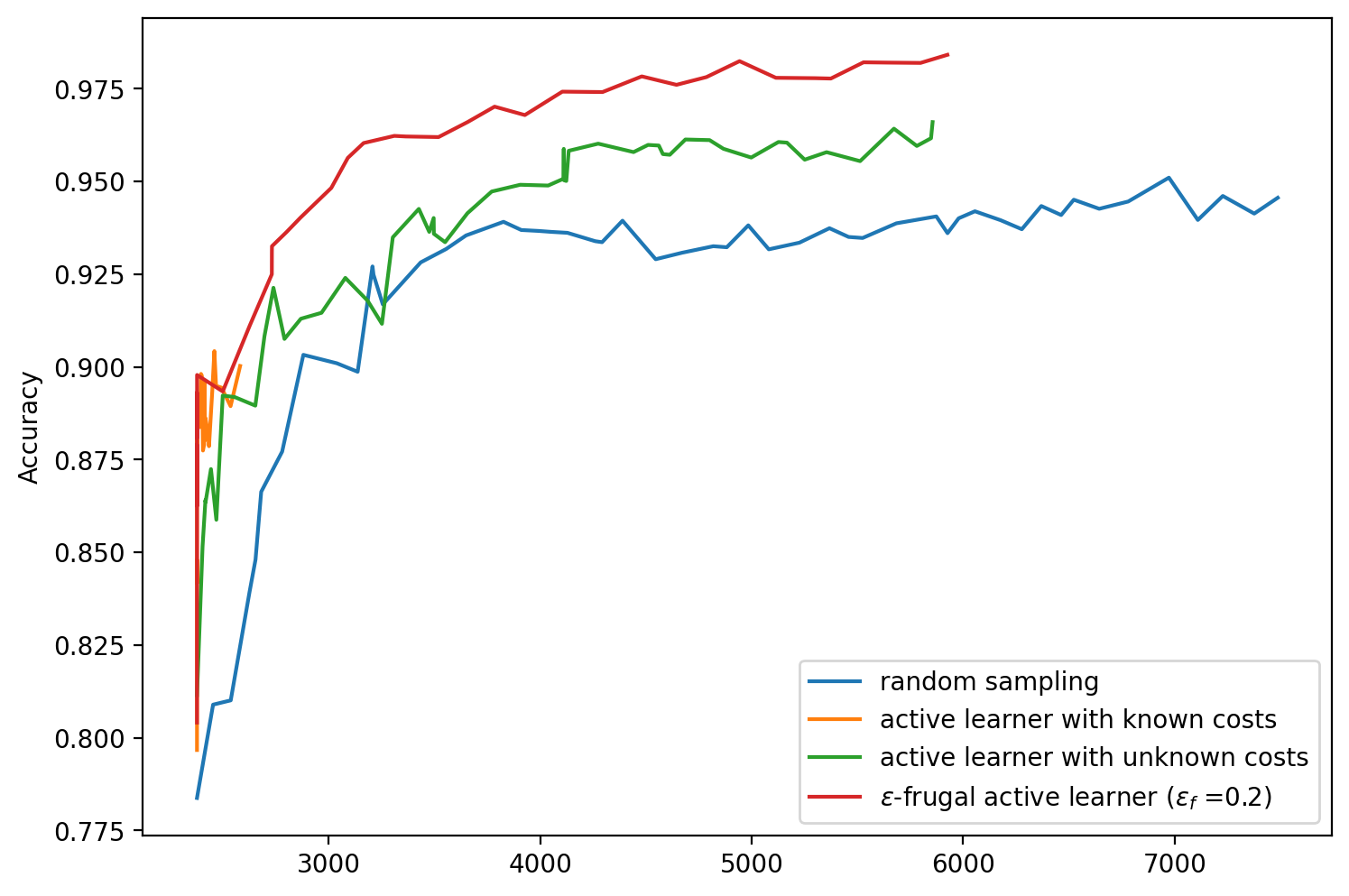}
	\caption{Comparison of the accuracy of training the model without using an active learner (random sampling), active learner with known costs, unknown costs, and $\epsilon$-frugal active learner (where $\epsilon$=0.2) with known costs. Note that the x-axis is now over the cumulative cost of sampling. We find that active learning both with unknown costs or frugal costs obtains high accuracy with lower overall sampling costs. Query selection is entropy-based for all active learners}
	\label{fig:comparingcostcalcs}
\end{figure}

In \autoref{fig:comparingcostcalcs}, we compare the three cost-aware active learners to random sampling. Here, we plot the learning curves against the overall costs for all sampling procedures, where sampling occurs over 50 iterations and with 2 samples queried for each one. Overall, we find that the random sampling has both the highest cumulative cost with lowest accuracy. The frugal active learner with known costs, where sampling is chosen purely to minimize costs and certainty has the lowest overall costs but significantly reduced accuracy compared to all other sampling methods. Both the $\epsilon$-frugal and the learner with unknown (but learned) costs have similarly low costs but much higher accuracy. The $\epsilon$-frugal learner has the highest accuracy of all learners (99.3\%).

These results show the benefit of introducing cost-awareness to an active learning framework. Not only are costs reduced, but reductions occur without any trade-off in accuracy. This is robust both to situations where costs are known or unknown at query stage and especially where costs are not uniform across the sample-space.

\section{Future Extensions}

In this work, we present an extension to the traditional active learning framework by introducing cost-awareness to the learner. We highlight two novel features, the concept of $\epsilon$-frugal learning and how the choice of the query selection measure can impact the structure of the sample cost-rank relationship. This active learning framework is straightforward to implement but can lead to significant savings in model exploration without a corresponding drop in accuracy. Furthermore, our active learner is robust to situations where costs are either known or unknown at query stage. 

We have chosen a simple active learner framework to demonstrate our extension and applied it to a well-known classification data-set. However, these restrictions are easily relaxed to allow for learners of more complicated models using less well-structured data. Furthermore, our methodology is not restricted to classification problems and is expected to perform just as well in other semi-supervised learning scenarios such as regression and inverse design. 

In the future, we will extend our active learner in several key ways. First, we will introduce a committee of learners in that several models are trained simultaneously on the test data and query-sampling will be a function of the disagreement between this committee of models. This provides interesting future research potential in both the influence of costs where committee ranking occurs and in designing committee's of learners that promote novel (weakly explored) regions of the sample space. 

Second, we will introduce causality into the active learning framework. This can be achieved by, for example, using models that explicitly account for causal relations such as generalized random forests \cite{athey2019generalized} or Bayesian optimization routines \cite{Ghahramani2015}. This focus on causal inference allows for interpretable learning that can be better understood by human operators. 

Finally, we will apply our active learner to a combination of simulation and experiment data, whereby feedback from experiments act as an Oracle for labeling samples and simulations of the experiment forms the model. This simulation-experiment learning framework allows for biological and physical relations to be tested to give significantly different types of causal information than that gained from statistical models, offering a new form of automated scientific discovery.

\section{Acknowledgments}
This research has received funding from the European Union’s Horizon 2020 research and innovation programme under grant agreement No 800983.\\

\bibliographystyle{./IEEEtran}
\bibliography{learning}

\end{document}